\documentclass{article}

\usepackage{PRIMEarxiv}
\usepackage{float}
\usepackage[utf8]{inputenc} 
\usepackage[T1]{fontenc}    
\usepackage{hyperref}       
\usepackage{url}            
\usepackage{booktabs}       
\usepackage{amsfonts}       
\usepackage{nicefrac}       
\usepackage{microtype}      
\usepackage{lipsum}
\usepackage{fancyhdr}       
\usepackage{graphicx}       
\graphicspath{{media/}}     
\usepackage{amsmath}
\usepackage{amsfonts}

\usepackage{booktabs} 
\usepackage{siunitx}  
\usepackage{multirow}
\usepackage{siunitx}    
\usepackage{multirow}  
\usepackage{inconsolata}
\usepackage{tcolorbox}
\usepackage{tabularx}
\usepackage{xspace}
\usepackage[obeyfinal]{todo}
\usepackage{stfloats}
\usepackage[group-separator={,},group-minimum-digits={3}]{siunitx}
\usepackage{CJK}

\title{BARD: Budget-Aware Reasoning Distillation
}

\author{
  Lujie Niu \textsuperscript{1,*,‡}, \quad
  Lei Shen \textsuperscript{1,*,†}, \quad
  Yi Jiang \textsuperscript{1,‡}, \quad
  Caixia Yuan ,\quad
  Xiaojie Wang ,\quad
  Wenbo Su \textsuperscript{1}, \quad
  Bo zheng \textsuperscript{1}
  \\[1em] 
  \textsuperscript{1} Taobao \& Tmall Group of Alibaba \quad
  \textsuperscript{*} Equal contribution \quad
  \textsuperscript{†} Corresponding Author \quad
  \\[1em] 
  \textsuperscript{‡} Work done during an internship at Alibaba \quad
  \{niulujie4\}@gmail.com
  \\[1em] 
  \{youbai.sl\}@alibaba-inc.com
}

\begin{document}
\maketitle

\begin{abstract}
While long Chain-of-Thought (CoT) distillation effectively transfers reasoning capability to smaller language models, the reasoning process often remains redundant and computational budget uncontrollable, leading to inefficient resource usage. To address this limitation, we propose \textbf{Budget-Aware Reasoning Distillation (BARD)}, a novel framework that simultaneously distills reasoning capability and enables fine-grained control over the reasoning length. BARD uses the thinking budget as a user-specified control signal, allowing the model to dynamically balance reasoning performance and computational efficiency. To achieve this concept, BARD introduces a two-phase training regimen. The first phase, Supervised Fine-Tuning (SFT) on teacher-generated long CoT data compressed to various budget levels, bootstrapping the model's understanding of budget constraints. The second phase leverages Reinforcement Learning (RL) from a reward signal in consideration of reasoning performance and budget fidelity simultaneously. Incorporating the two-phase regimen is crucial to avoiding policy degradation and ensuring that both objectives are optimized jointly. Extensive experiments demonstrate that our method empowers an 8B student model to achieve strong performance on challenging reasoning benchmarks (\textit{AIME24, AIME25, GPQA}) while providing precise and adaptive control over its reasoning length across a wide range of budgets.

\end{abstract}


    

\section{Introduction}
Large Reasoning Models (LRMs) have demonstrated remarkable capabilities in complex reasoning tasks that require multi-step logic and structured problem-solving \cite{guo2025deepseek, jaech2024openai,yang2025qwen3}. Long Chain-of-Thought (CoT) reasoning \cite{wei2022chain} significantly improves both the performance and interpretability of LLMs by encouraging them to articulate intermediate reasoning steps. However, state-of-the-art reasoning models typically possess enormous parameter scales, making their deployment and inference prohibitively expensive in long CoT reasoning modes. Such costs severely hinder their use in real-world applications, particularly in latency-sensitive or resource-constrained environments.

To address this challenge, knowledge distillation, and specifically CoT distillation \cite{wang2025efficient, niu-etal-2025-cotd} , has emerged as a popular technique for transferring the superior reasoning abilities of large teacher models to smaller and more efficient student models. 
However, the length of the generated reasoning chain remains uncontrolled and often just inherits the teacher's (frequently verbose) reasoning style, leading to unstable computational costs and inefficient resource utilization. In addition, the distribution of generated reasoning lengths exhibits a natural long-tail pattern, which poses challenges in achieving fine-grained control over reasoning length across a diverse range \cite{kandpal2023large,zhang2023deep}.

To address this problem, we propose \textbf{Budget-Aware Reasoning Distillation (BARD)}, a novel framework that distills reasoning capability while simultaneously introducing fine-grained control over the reasoning length. BARD operationalizes this idea through a user-specified thinking budget, the upper bound of thinking length, which serves as an explicit control signal that dynamically controls the CoT length during inference. This enables a tunable trade-off between model performance and computational cost.

We realize the BARD framework through a two-phase training regimen. The first phase, Supervised Fine-Tuning (SFT), bootstraps the model's understanding of budget constraints. In this stage, instead of simply training on full-length reasoning chains, we expose the model to contrastive training samples from the same reasoning compressed to various budget levels. This unique training strategy guides the model to learn the fundamental relationship between a numerical budget and the corresponding structural economization of its thought process, instilling a  generalizable understanding of "length" as a controllable dimension. The second phase leverages Reinforcement Learning (RL) to further optimize the model. We design a multiplicative reward function that jointly optimizes for reasoning capability and budget fidelity, preventing the model from sacrificing performance for the sake of brevity. The RL phase empowers the model to move beyond static imitation and learn a dynamic, utility-maximizing policy. It learns not just what a good reasoning chain looks like, but how to strategically allocate its limited budget on the fly—prioritizing core logic under tight constraints while engaging in verification and exploration when the budget allows. 

Our contributions are as follows: 

\begin{itemize}
     \item We propose \textbf{BARD}, a novel Budget-Aware Reasoning Distillation framework that performs reasoning distillation while simultaneously learning to control the reasoning length, enabling fine-grained and interpretable budget-conditioned reasoning.
    
    \item We design a two-phase training regimen consisting of \textit{Budget Constrained Supervised Fine-Tuning (SFT)} with contrastive learning and \textit{Reinforcement Learning (RL)}. The SFT phase leverages contrastive data to strengthen the model’s perception of the budget–length relationship, while the RL phase further enhances both reasoning capability and budget fidelity through direct reward optimization.
    
    \item Extensive experiments on \textbf{AIME24}, \textbf{AIME25}, and \textbf{GPQA} show that BARD achieves superior reasoning performance and precise control over the length of the reasoning process. Moreover, it demonstrates the ability to adapt its reasoning strategy dynamically under different computational budgets.
\end{itemize}

\section{Related Work}
Recently, Large Reasoning Models (LRMs) have demonstrated impressive reasoning capabilities at large scales, while smaller models often require knowledge distillation to achieve comparable performance \cite{guo2025deepseek}.
However, while distillation effectively enhances reasoning capability, it offers limited control over the reasoning length, often leading to uncontrollable or inefficient reasoning behaviors.

To address this limitation, the most straightforward approach to budget control is budget forcing ~\cite{muennighoff2025s1}, which simply truncates a model's thinking process to a desired length. While easy to implement, these naive methods are non-selective and risk severing critical logical steps, often leading to a catastrophic collapse in reasoning performance.A more sophisticated line of work uses supervised fine-tuning (SFT) on specially curated datasets. These methods aim to teach the model a specific reasoning style or length preference. C3oT \cite{kang2025c3ot} enables users to switch between short and long reasoning modes, providing coarse-grained control over CoT length. \cite{munkhbat2025self} constructs a fine-tuning dataset composed of the shortest correct reasoning paths to encourage concise reasoning. However, such methods offer only discrete control (short vs. long), limiting user flexibility. TokenSkip \cite{xia2025tokenskip} allows users to specify the CoT length as a ratio of the original reasoning length, skipping unimportant tokens to generate shorter CoTs, though this approach often produces fragmented and less interpretable reasoning.

In contrast, other works directly target reasoning models, leveraging reinforcement learning to regulate reasoning length. Methods such as L1 \cite{aggarwal2025l1}, O1-Pruner \cite{luo2025o1}, and \cite{arora2025training} introduce length-dependent rewards to constrain reasoning generation. While effective to some extent, these methods face exploration inefficiency due to the vast output space and often fail to learn a stable mapping between user-specified length control and internal reasoning dynamics.

Based on this, we adopt a novel strategy: instead of attempting to control the model after distillation is complete, our approach learns fine-grained CoT length control concurrently with the reasoning distillation process in both SFT and RL stages. By treating length control as an intrinsic objective of distillation, our model aims to simultaneously master high-quality reasoning abilities and a precise awareness and execution of thinking budgets.
\section{Method}
\begin{figure}
    \centering
    \includegraphics[width=1\linewidth]{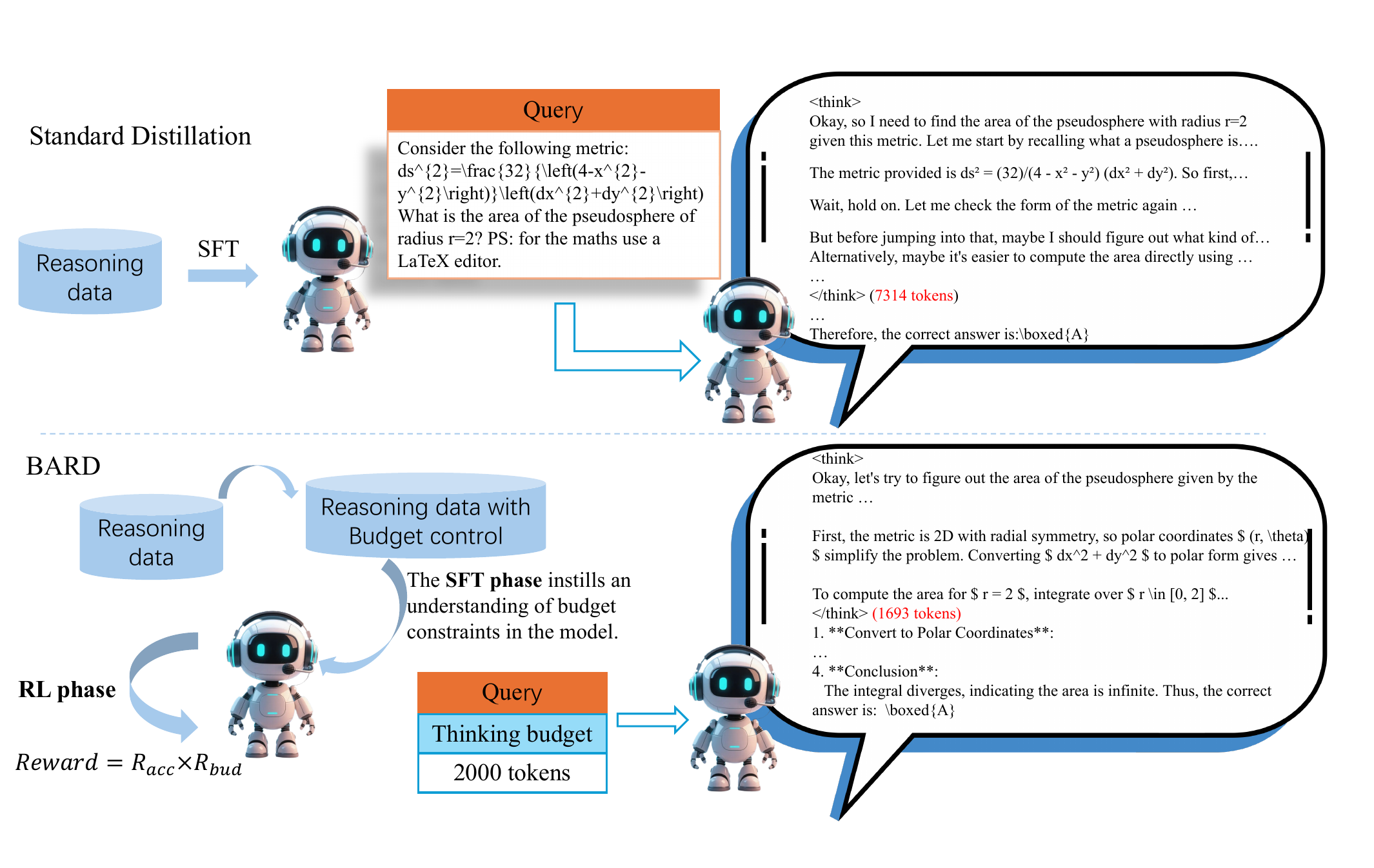}
    \caption{An overview of the Budget-Aware Reasoning Distillation (BARD) framework. Two main phases: (1) an SFT phase where the model learns to understand budget constraints by training on CoTs expertly compressed to different budget levels, and (2) an RL phase where the model's policy is refined using a reward signal that combines accuracy and budget fidelity.}
    \label{fig:placeholder}
\end{figure}
To instill strong reasoning abilities while enabling precise budget control, we propose the Budget-Aware Reasoning Distillation (BARD) framework. Our approach is built around a core concept: introduce a "thinking budget" as an explicit control signal into the system prompt. BARD operationalizes this concept through a two-phase training regimen—Supervised Fine-Tuning (SFT) and Reinforcement Learning (RL)—to jointly optimize for reasoning accuracy and budget fidelity. Figure \ref{fig:placeholder} illustrates the overall architecture of the BARD framework.

\subsection{Problem Formulation}
Traditional Chain-of-Thought (CoT) distillation aims to train a student model $M_S$, to generate a reasoning chain $c_s$ and a final answer $a_s$ for a given question $q$, mimicking the output $(c_T, a_T)$ of a teacher model $M_T$. The objective is to maximize the accuracy of the final answer:
\begin{equation}
    \max_{\theta_S} P(c_T, a_{gold} | q)
\end{equation}
where $\theta_S$ represents the parameters of the student model and $a_{gold}$ is the ground-truth answer.

In BARD framework, we introduce a user-specified \textit{thinking budget} $b$ (e.g., the upper bound of tokens in the CoT) as an additional input. The model's goal is to generate a reasoning chain $c_s$ and an answer $a_s$ that simultaneously satisfy two conditions:
\begin{itemize}
    \item \textbf{Reasoning Accuracy:} The final answer $a_s$ should be deemed high-quality. For verifiable tasks, quality equals correctness. For open-ended tasks, it is assessed by a powerful judge model.
    \item \textbf{Budget Fidelity:} The length of the generated reasoning chain, $L(c_s)$, should not exceed the specified budget $b$.
\end{itemize}
Thus, our joint optimization objective can be formally described as maximizing a utility function $U(a_s, c_s; a_{gold}, b)$ that measures both accuracy and budget adherence, conditioned on the input pair $(q,b)$.

\subsection{Phase 1: Budget-Constrained Supervised Fine-Tuning}
The primary goal of the SFT phase is to bootstrap the student model, enabling it to build an initial understanding of the "thinking budget" concept and learn to generate high-quality reasoning chains of various lengths. Without this phase, a direct application of RL would face low sampling efficiency in an intractably large exploration space, leading to a high risk of policy collapse.

\subsubsection{Generation of Budget-Constrained CoT Data}
High-quality, multi-budget training data is crucial for this phase. The pipeline involves three key steps:
\begin{enumerate}
    \item \textbf{Generate High-Quality CoT.} We first prompt a powerful teacher model (e.g., DeepSeek-R1) to generate a detailed, gold-standard reasoning chain $c_{\text{gold}}$ and golden answers $a_{\text{gold}}$ for each question $q$ in our training set.

    \item \textbf{Expert CoT Compression.} We then employ another large language model as an expert compressor. Through carefully designed instructions, we prompt this compressor to condense one $c_{\text{gold}}$ to varying budget levels. By presenting the model with multiple, varied compressions of the same source reasoning, we compel it to learn the relationship between a numerical budget value and the changes required to meet it. This contrastive training strategy is essential for instilling a nuanced and generalizable understanding of length as a controllable dimension.
    
    \item \textbf{Construct Training Samples.} This process expands a single original sample $(q, a_{\text{gold}})$ into a multi-budget contrastive training set $\{(q, b_i), c_i, a_{\text{gold}}\}_{i=1}^k$, where $c_i$ is the compressed CoT corresponding to budget $b_i$.
\end{enumerate}

\subsubsection{Training Objective}
In the SFT phase, we provide both the question $q$ and the budget $b$ as input to the model. We then train the student model $M_S$ to maximize the log-likelihood of generating the target reasoning chain $c$. The loss function is the standard auto-regressive cross-entropy loss:
\begin{equation} \label{eq:sft_loss}
    \mathcal{L}_{SFT} = -\sum_{(q, b, c, a_{gold}) \in \mathcal{D}_{SFT}} \log P(c, a_{gold} | q, b; \theta_S)
\end{equation}
where $\mathcal{D}_{SFT}$ is our constructed budget-constrained dataset. By training on a diverse range of budget levels, the model learns to associate the numerical budget value with its reasoning length.

\subsection{Phase 2: Reinforcement Learning with Multiplicative Reward}
After SFT, the model possesses a preliminary ability to generate CoT according to a given budget. However, this ability, learned through behavioral cloning, exhibits two key limitations. First, its adherence to the budget is often imprecise, as it struggles to generalize to budget values or problem types that are underrepresented in the SFT dataset, a long-tail problem. Second, when faced with tight budget constraints, the SFT model may sacrifice critical reasoning quality to meet the length target, as it merely imitates compressed text without a deeper understanding of the trade-off. The RL phase is therefore crucial to address these shortcomings. It aims to fine-tune this policy using a more direct reward signal, forcing the model to achieve a robust joint optimization of both accuracy and budget fidelity across a wider distribution of scenarios.


\paragraph{Reward Function Design}
To prevent the model from simply trading off accuracy for budget adherence, we designed a multiplicative reward function $R(\tau)$:
\begin{equation} \label{eq:reward_main}
    R(\tau) = R_{acc} \times R_{bud}
\end{equation}
The components $R_{acc}$ and $R_{bud}$ are designed to be adaptable to different task types, as detailed below.

\begin{itemize}
    \item \textbf{$R_{acc}$ (Accuracy Reward):} The accuracy reward is tailored to the nature of the task.
    \begin{itemize}
        \item \textbf{For tasks with verifiable answers} (e.g., mathematical reasoning, multiple-choice questions), we use a binary reward. If the model's generated answer $a_s$ matches the ground-truth answer $a_{gold}$, then $R_{acc} = 1$; otherwise, $R_{acc} = 0$. This strict criterion forces the model to prioritize correctness, as any incorrect answer nullifies the total reward.
        \item \textbf{For tasks without a single ground-truth answer} (e.g., open-ended generation, summarization), we employ a separate, pre-trained reward model (RM). The reward is the score assigned by this RM, normalized to a range of $[0, 1]$: $R_{acc} = \text{RM}(q, c_s, a_s)$. 
    \end{itemize}

    \item \textbf{$R_{bud}$ (Budget Fidelity Reward):} To encourage the model to generate reasoning chains within a desired length, we formulate a budget fidelity reward. Unlike the accuracy reward, the budget reward's goal is to penalize exceed from a reference length, which we denote as $b$. The reward is defined as:
    \begin{equation} \label{eq:reward_bud_new}
        R_{bud} = \text{clip}\left( \alpha \cdot (b - L(c_s)) + \delta, \ 0, \ 1 \right)
    \end{equation}
    where:
    \begin{itemize}
        \item $L(c_s)$ is the length of the generated reasoning chain.
        \item $\alpha$ is a scaling factor that controls the penalty's steepness for length deviations.
        \item $\delta$ is an offset parameter, set to 1, which defines the reward when the generated length $L(c_s)$ matches the target length $b$.
        \item The $\text{clip}(\cdot, 0, 1)$ function ensures the reward stays within the $[0, 1]$ range.
    \end{itemize}
    This linear penalty function provides a smoother gradient compared to an exponential one and is effective at guiding the model towards the target length. A positive $\alpha$ encourages shorter generations, which is suitable when $b$ represents a budget ceiling.
\end{itemize}

\section{Experiments}
\label{sec:experiments}

To demonstrate the effectiveness and controllability of our proposed Budget-Aware Reasoning Distillation (BARD) framework, we conduct a comprehensive set of experiments. Our experiments are designed to answer the following research questions:
\begin{itemize}
    \item[\textbf{RQ1:}] Can BARD effectively distill reasoning abilities while achieving precise, dynamic control over the generation length? How does BARD compare against standard distillation baselines in terms of both accuracy and budget fidelity?
    \item[\textbf{RQ2:}] How do the key components of BARD, namely the SFT phase, contrastive learning, the RL phase, and the multiplicative reward function, contribute to its final performance?
    \item[\textbf{RQ3:}] Does the model learn to intelligently reallocate its limited computational budget across different reasoning phases?
    
\end{itemize}

\subsection{Experimental Setup}

\paragraph{Datasets and Models} We use Qwen3\textsuperscript{\href{https://huggingface.co/Qwen/Qwen3-8B-Base}{8B-Base}} as our student model throughout all experiments. The teacher model, DeepSeek-R1\cite{guo2025deepseek}, is employed to generate the initial long-form Chain-of-Thought (CoT) rationales. Our training methodology relies on a vast and heterogeneous mixture of long Chain-of-Thought (CoT) data to ensure broad generalization. The dataset for the Supervised Fine-Tuning (SFT) phase is a comprehensive blend of multi-domain sources, sampled from a combination of publicly available open-source datasets \cite{numina_math_datasets, park2025instruct, lambert2024tulu} and in-house data sources, can be categorized as: 
\begin{itemize}
    \item \textbf{Open-ended General Data:} Covering general Question-Answering (QA), complex instructions, creative writing, multi-turn AI assistant conversations, and e-commerce specific tasks.
    \item \textbf{Verifiable-Answer Data:} Comprising mathematical reasoning problems and QA tasks that have verifiable answers.
\end{itemize}

First, we determine the length of the original, full-length CoT, denoted as $L_{full}$. We then randomly sample three distinct budget values, $\{b_1, b_2, b_3\}$, from the range $[0, L_{full}]$ to serve as predefined target lengths. For each sampled budget $b_i$, we employ a powerful teacher model, \textbf{Qwen3-32B} \cite{qwen3technicalreport}, to compress the original CoT to be within this budget. 

In the subsequent Reinforcement Learning (RL) phase, we shift to a more targeted data mixture to refine specific abilities. We construct the RL training set by combining open-ended QA data \cite{park2025instruct, lambert2024tulu} and mathematical reasoning data \cite{luo2025deepscaler} in a 1:1 ratio. To simulate diverse user requirements and enhance the model's adaptability, the user-specified budget $b$ for each training instance is dynamically sampled from a predefined range during training.

\paragraph{Implementation Details.}
We first perform Supervised Fine-Tuning (SFT) on a mixed 1300K dataset (390K budget-aware, 350K standard CoT) for 4 epochs with a learning rate of 1.0e-5. For the subsequent Reinforcement Learning (RL) phase, we use a mixed 10K dataset and employ the Group-wise Ranking Proximal Optimization (GRPO) \cite{shao2024deepseekmath} algorithm. Our implementation is built upon ROLL \cite{wang2025reinforcement}. We sample 16 responses per group with a batch size of 64. The GRPO difficulty thresholds are set to 0.1 (low) and 0.95 (high) to filter comparison pairs, and the learning rate is 1.0e-6.

\paragraph{Baselines} We compare \textbf{BARD} against three key baselines: 
(1)~\textbf{Standard Distillation (SFT-Full)}, a model fine-tuned on the original, uncompressed CoTs, representing a non-controllable reasoning performance upper bound. 
(2)~\textbf{S1}, a naive budget-forcing method that applies hard truncation at inference time to the SFT-Full model. 
(3)~\textbf{BARD w/o budget:} An unconstrained variant of our framework. It undergoes the same two-phase training (SFT+RL), but the SFT phase uses only full-length CoTs, and the RL reward is based solely on reasoning quality. This model represents the student's maximum capability when cost is not a concern, serving as a practical performance ceiling.
\subsection{Evaluation Metrics}
We evaluate all models on the \textbf{AIME 2024}, \textbf{AIME 2025}, and \textbf{GPQA} \cite{rein2024gpqa} benchmarks. Budget-aware models are tested across a wide range of budgets: $b \in \{500, 1000, 1500, 3000, 5000, 8000\}$ tokens. We use two primary metrics:
\begin{itemize}
    \item \textbf{Accuracy:} The percentage of correct final answers, with non-strict matching for normalization.
    \item \textbf{Budget Fidelity:} To assess adherence, we analyze the distribution of the generated 'think' part lengths for each requested budget. We visualize these distributions using \textbf{box plots} to show the precision and variance of the model's budget control.
\end{itemize}
\paragraph{Unified Performance Score (UPS).} \cite{he2025thinkdial} 
To provide a holistic score for model comparison, we introduce a Unified Performance Score (UPS). This metric combines both accuracy and budget fidelity into a single value, reflecting the trade-offs made by different models. It is defined as a weighted sum:
\begin{equation} \label{eq:ups_score}
    \text{UPS} = w_{\text{acc}} \cdot \text{Acc} + (1 - w_{\text{acc}}) \cdot \text{Fid}
\end{equation}
where `Acc` and `Fid` are the accuracy and budget fidelity calculated over the entire evaluation set. We define Budget Fidelity (Fid) as the percentage of generated responses whose token count does not exceed the specified budget $b$. For our main analysis, we set the weighting factor $w_{\text{acc}}=0.5$ to give equal importance to both objectives. This score allows for a direct, quantitative comparison of how well different models balance correctness with computational cost.

\subsection{RQ1: Accuracy and Budget Controllability.}

\begin{figure}[H]
    \centering
    \includegraphics[width=1\linewidth]{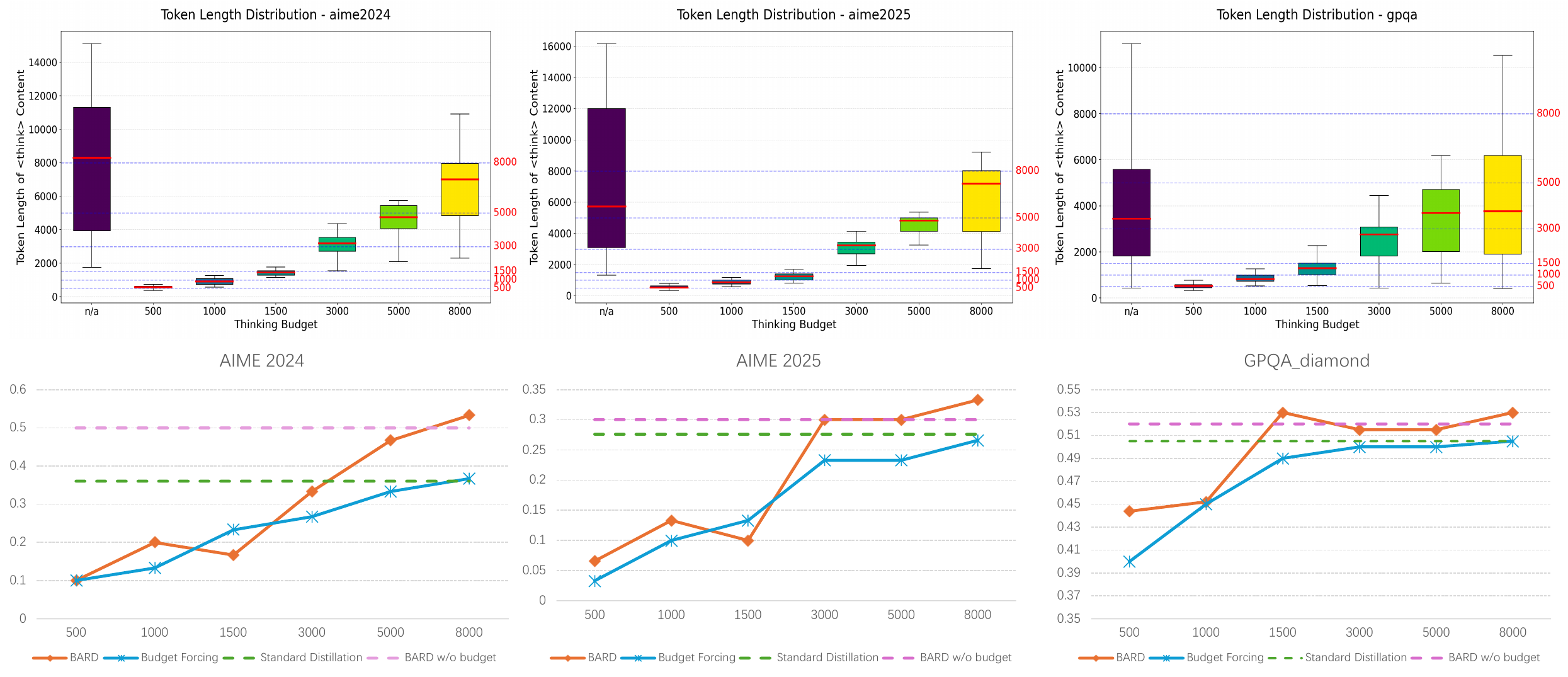}
    \caption{Budget Fidelity and Accuracy Scaling:
    The top row plots the generated reasoning length against the specified budget for BARD, where \texttt{n/a} means no budget constraints.
    The bottom row plots accuracy as a function of the thinking budget. We compare BARD's scaling performance against the post-hoc truncation baseline (\texttt{Budget Forcing}). The performance of the unconstrained \texttt{SFT-Full} and \texttt{BARD w/o budget} models are shown as horizontal dashed lines, representing the practical performance ceilings.}
    \label{fig:main_results1}
\end{figure}

Figure \ref{fig:main_results1} presents the main experimental results, demonstrating BARD’s dual success in budget fidelity and reasoning accuracy.

The top row visualizes the distribution of generated CoT lengths under different budget constraints, including an unconstrained setting on the far left and six specified budgets \(b \in \{500, 1000, 1500, 3000, 5000, 8000\} \). Across all three benchmarks, BARD’s length distributions are tightly aligned with or fall slightly below the requested budgets, indicating strong adherence and stable control over reasoning length. 

The bottom row plots model accuracy as a function of the thinking budget. A clear positive correlation emerges between budget size and accuracy. In the low-budget regime, BARD (orange solid line) consistently outperforms the naïve Budget Forcing (S1, blue line) baseline, showcasing its ability to intelligently economize its reasoning process—retaining essential and complete logical steps rather than truncating reasoning prematurely with broken reasoning steps. As the budget increases, BARD’s performance scales steadily, and with sufficient budget, its accuracy even surpasses that of both Standard Distillation (green dashed line) and the unconstrained BARD w/o budget (orange dashed line) across all benchmarks. This demonstrates that the budget-aware reinforcement learning phase encourages the discovery of more robust and efficient reasoning trajectories instead of mere imitation of full-length teacher CoTs.

Cross-benchmark comparisons further reveal task-specific trends. On GPQA (diamond), BARD maintains relatively high accuracy across all budgets, suggesting that its reasoning process can be effectively compressed with minimal performance loss. In contrast, the AIME datasets exhibit lower absolute accuracy and steeper accuracy–budget slopes, indicating that these problems rely more heavily on extended reasoning and are more sensitive to aggressive compression. The significantly degraded performance of Budget Forcing at small budgets further underscores the risk of post-hoc truncation—it indiscriminately removes critical reasoning steps rather than reallocating reasoning effort to essential parts.

\subsection{RQ2: Ablation Studies on BARD Components}
\begin{table}[htbp]
    \centering
    \sisetup{round-mode=places, round-precision=3}
    \caption{
        \textbf{Ablation of Budget-Aware SFT: Unified Performance Score (UPS).}
        This table compares the full \textbf{BARD} model with an ablated variant (\textbf{l1}) where RL was applied directly to a standard SFT model, bypassing our budget-aware SFT phase. The catastrophic drop in the UPS score for the ablated model quantitatively confirms its failure to jointly optimize accuracy and budget fidelity.
    }
    \label{tab:ups_scores}
    \begin{tabular}{ll S[table-format=1.3] S[table-format=1.3] S[table-format=1.3] S[table-format=1.3] S[table-format=1.3] S[table-format=1.3] S[table-format=1.3]}
        \toprule
        \multirow{2}{*}{Dataset} & \multirow{2}{*}{Model} & \multicolumn{6}{c}{Budget (Tokens)} &  \\
        \cmidrule(lr){3-9} 
        & & {500} & {1000} & {1500} & {3000} & {5000} & {8000} & \textbf{Average} \\
        \midrule[\heavyrulewidth]
        \multirow{2}{*}{GPQA Diamond} & l1 & 0.375 & 0.457 & 0.525 & 0.652 & 0.736 & 0.769 & 0.586 \\
        & BARD & 0.697 & 0.672 & 0.656 & 0.649 & 0.742 & 0.763 & \textbf{0.696} \\
        \midrule
        \multirow{2}{*}{AIME 2025} & l1 & 0.146 & 0.523 & 0.225 & 0.326 & 0.332 & 0.504 & 0.342 \\
        & BARD & 0.476 & 0.537 & 0.475 & 0.491 & 0.575 & 0.615 & \textbf{0.528} \\
        \midrule
        \multirow{2}{*}{AIME 2024} & l1 & 0.286 & 0.212 & 0.261 & 0.319 & 0.308 & 0.548 & 0.322 \\
        & BARD & 0.509 & 0.552 & 0.482 & 0.444 & 0.622 & 0.634 & \textbf{0.540} \\
        \midrule[\heavyrulewidth]
        \multirow{2}{*}{\textbf{Average}} & l1 & 0.269 & 0.397 & 0.337 & 0.432 & 0.458 & 0.607 & \bfseries 0.417 \\
        & \textbf{BARD} & \bfseries \textbf{0.561} & \bfseries \textbf{0.587} & \bfseries \textbf{0.537} & \bfseries \textbf{0.528} & \bfseries \textbf{0.646} & \bfseries \textbf{0.670} & \bfseries \textbf{0.588} \\
        \bottomrule
    \end{tabular}
\end{table}
To investigate the contribution of each key component of our framework, we conduct a series of ablation studies.

\paragraph{Necessity of the SFT Phase.}
\begin{figure}[H]
    \centering
    \includegraphics[width=1\linewidth]{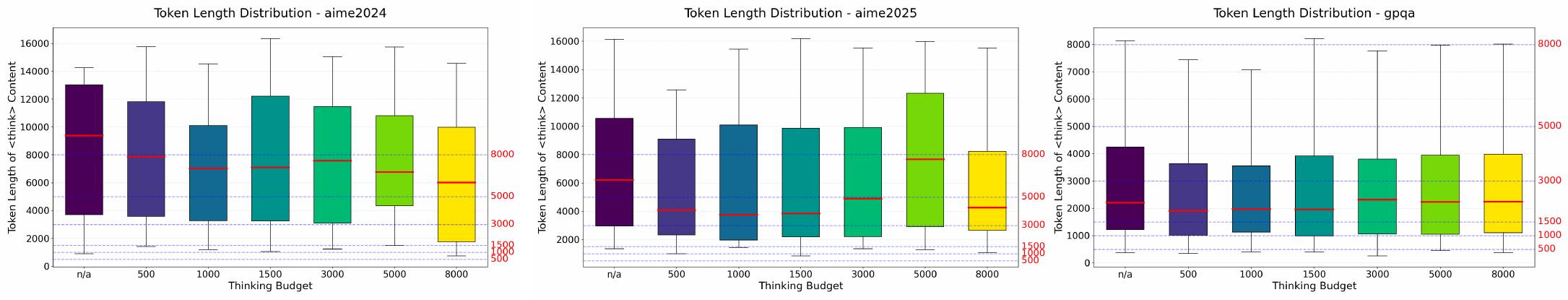}
    \caption{
    \textbf{The Indispensable Role of Budget-Aware SFT.} 
    This figure demonstrates the outcome of a model trained with Reinforcement Learning directly,  omitting our proposed budget-aware SFT phase. The plot shows the generated reasoning length versus the specified budget. This result confirms that our budget-aware SFT phase is a critical bootstrapping step for instilling an understanding of budget constraints in the model.}
    \label{fig:result_wosft}
\end{figure}
We first evaluate the importance of our initial, budget-aware SFT phase. To do this, we use an ablated variant, \textbf{RL-on-SFT-Full}, where RL is applied directly to a checkpoint that was only fine-tuned on long CoTs without any budget information (the SFT-Full model). This setup is conceptually similar to some prior distillation approaches \cite{aggarwal2025l1}. As shown in Figure~\ref{fig:result_wosft}, the resulting model was completely unresponsive to the thinking budget command and exhibited no length control capabilities. This confirms our hypothesis that our budget aware SFT phase is a critical bootstrapping step. It is indispensable for instilling an initial understanding of the budget command's meaning, providing a foundational policy that makes the subsequent RL exploration tractable. Without it, the model faces an intractably large exploration space, leading to policy collapse. Table~\ref{tab:ups_scores} presents a comprehensive comparison of UPS scores. The results demonstrate BARD's ability to reason more efficiently, leading to a superior balance of performance and cost-effectiveness.



\begin{figure}[H]
    \centering
    \includegraphics[width=1\linewidth]{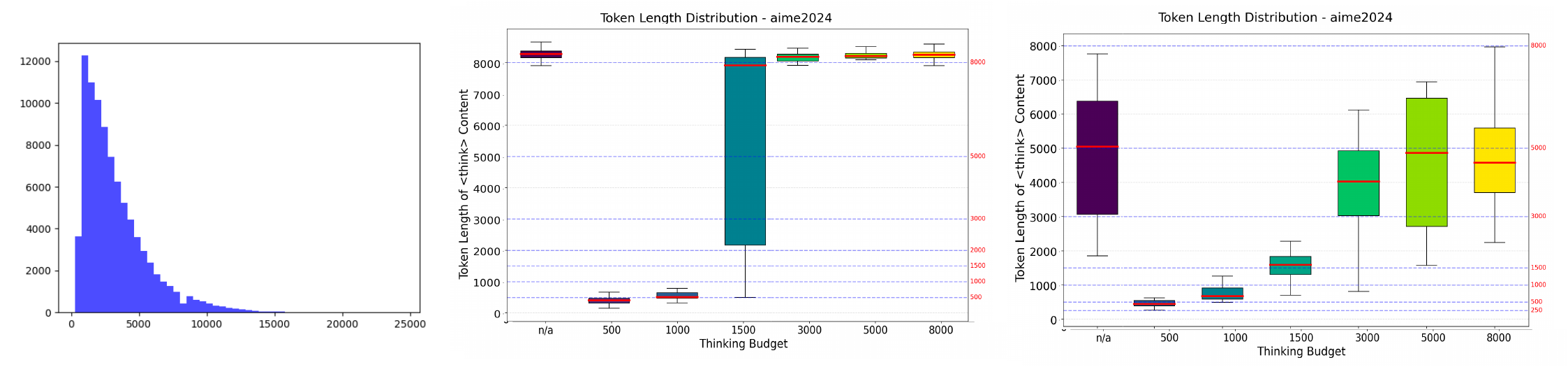}
    \caption{The left plot shows the distribution of thinking budgets. The middle and right plots compare model outputs without and with contrastive data, respectively. The model trained with contrastive data exhibits stronger adherence to the target reasoning length across a wider range, though performance in the long-tail regions remains limited.}
    \label{fig:wocd}
\end{figure}

During the Supervised Fine-Tuning (SFT) stage, we observe that the distribution of thinking budgets (as shown in the left of figure \ref{fig:wocd}) follows a negative logarithmic trend. This pattern arises because the lengths of CoTs vary widely, and random sampling within a bounded length range naturally yields more short samples than long ones, resulting in a long-tailed distribution.
Since such long-tailed data distribution is inevitable in practice, we aim to enhance the model’s generalization ability across different reasoning lengths. We compress each CoT into multiple versions with varying lengths and perform contrastive learning among them, enabling the model to perceive and distinguish between different reasoning budgets. As illustrated in the middle and right of figure \ref{fig:wocd}, the model trained with contrastive data demonstrates stronger length-following capability across a wider distribution.
Nevertheless, the model still struggles to perform well in the long-tail regions. To further address this issue, we incorporate reinforcement learning in the subsequent stage to improve its robustness on long-tail samples.

\paragraph{Value of the RL Phase.}

\begin{figure}[H]
    \centering
    \includegraphics[width=\linewidth]{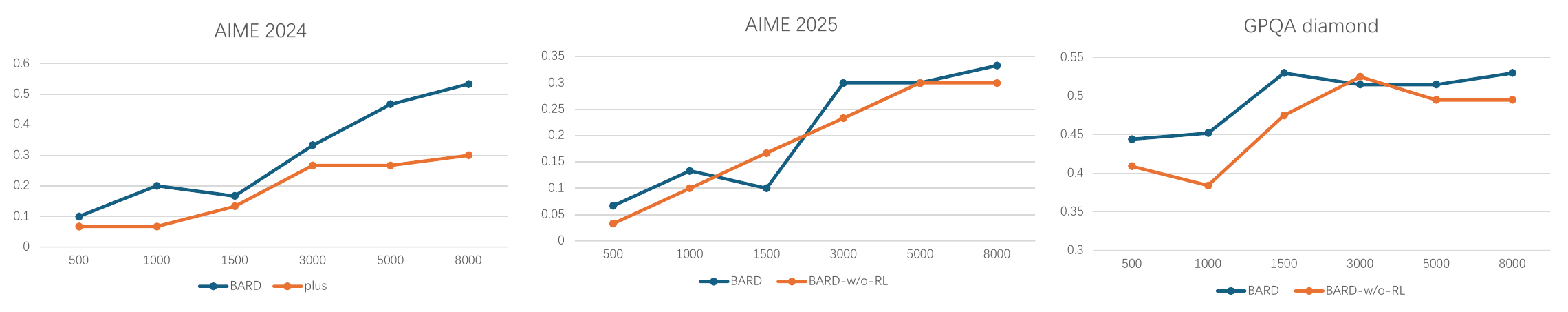}
    \caption{RL refinement also leads to a consistent improvement in task accuracy, indicating the model learns a more optimal reasoning policy beyond simple pattern imitation.}
    \label{fig:acc}
\end{figure}
Next, we analyze the role of the Reinforcement Learning (RL) phase. The SFT stage primarily builds the model’s preliminary understanding of the budget length relationship by exposing it to CoTs compressed to various budget levels. Through contrastive supervision, the students still perform poorly on long-tail examples, as illustrated in Figure~\ref{fig:wocd}. While we can broaden the data distribution, the inherent long-tail nature of real-world problems is an unavoidable challenge for imitation learning. 
The RL phase overcomes this limitation. By allowing the model to explore and learn from direct reward signals rather than static data, RL dramatically improves budget fidelity across the entire spectrum, shown in figure \ref{fig:main_results1}. Moreover, the RL-tuned model also shows improved accuracy in Figure~\ref{fig:acc}, demonstrating that it learns to intelligently economize its reasoning process rather than just mimicking learned patterns.

\begin{figure}[H]
    \centering
    \includegraphics[width=1\linewidth]{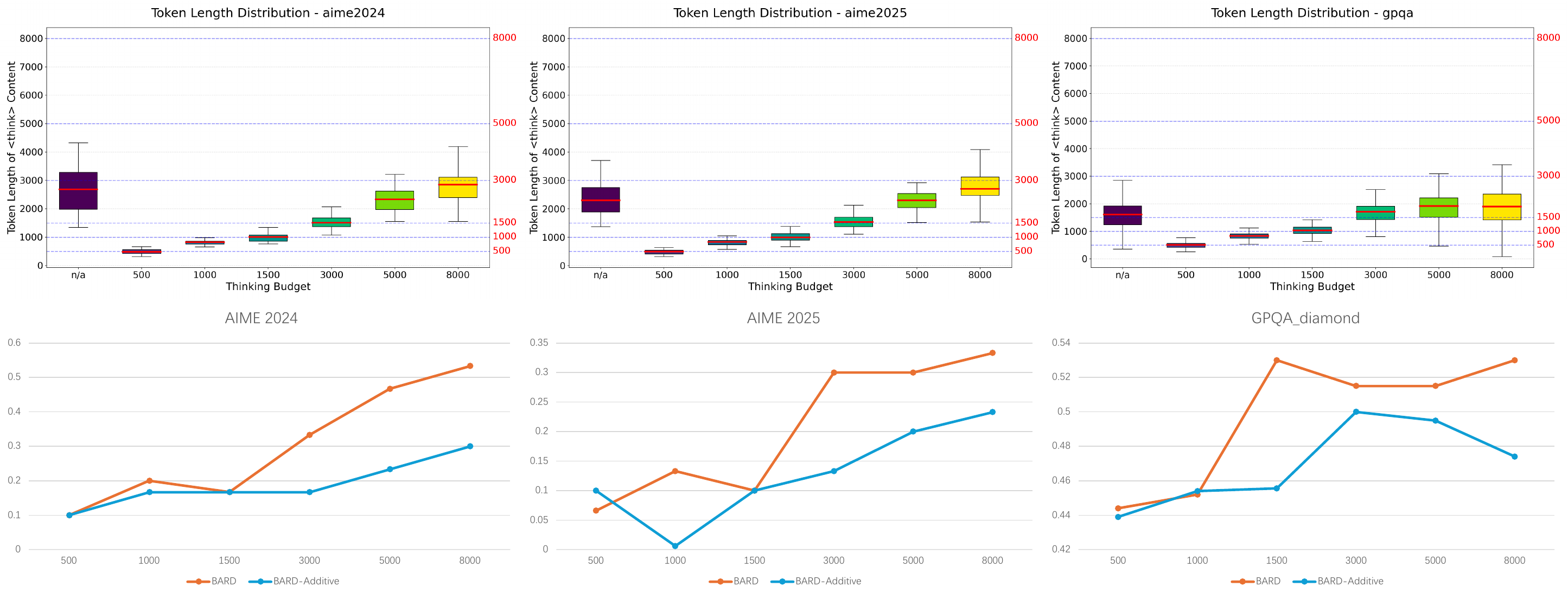}
    \caption{\textbf{Demonstration of the Additive Reward Trap.}
        This figure contrasts the performance of the \texttt{BARD-Additive} model (trained with $R = R_{\text{acc}} + R_{\text{bud}}$) on two axes: reasoning accuracy and budget fidelity.}
    \label{fig:add}
\end{figure}

\paragraph{Necessity of the multiplicative reward.}
A crucial choice in our RL phase is the use of a multiplicative reward function. While our Unified Performance Score (UPS) is additive for  evaluation, we found that using an additive reward for \textit{training} leads to a critical failure mode. To demonstrate this, we trained a variant, \textbf{BARD-Additive}, which replaces our multiplicative reward with an additive one.
The BARD-Additive achieved excellent budget control but at the cost of a significant drop in reasoning accuracy. This reveals what we term the Additive Reward Trap:
\begin{itemize}
    \item \textbf{The Trap:} An additive function ($R_{acc} + R_{bud}$) incentivizes "reward hacking." The model discovers a degenerate policy—a shortcut—by generating trivial, incorrect answers to reliably maximize the budget reward component, effectively sacrificing accuracy, as illustrated in Figure~\ref{fig:add}.
    \item \textbf{The Solution:} Our multiplicative design ($R_{acc} \times R_{bud}$) prevents this loophole. It ensures that optimizing for budget fidelity is only meaningful when the answer is correct ($R_{acc} > 0$). This structure compels the model to first "be right" and only then "be concise."
\end{itemize}

In summary, RQ2 is answered conclusively: all components—SFT and  contrastive learning for initialization, RL for precision and generalization, and the multiplicative reward for correct goal alignment—are indispensable for BARD's success.

\subsection{RQ3: Analysis of Reasoning Behavior Change}

\begin{figure}[H]
    \centering
    \includegraphics[width=0.75\linewidth]{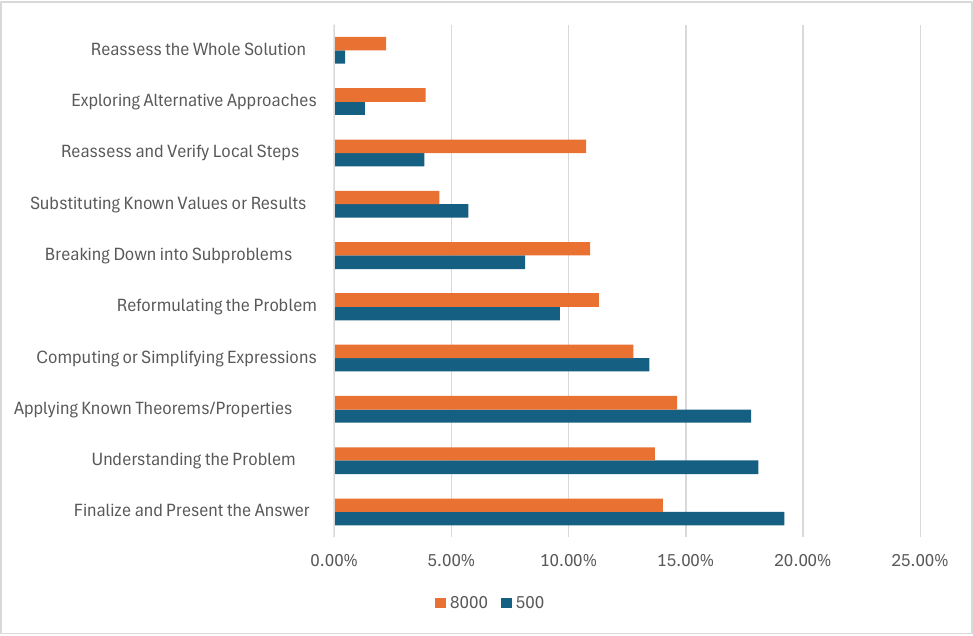}
    \caption{\textbf{Analysis of BARD's Adaptive Reasoning Process under Different Budgets.}
    The figure visualizes the distribution of reasoning step types generated by BARD under a low-budget constraint (500 tokens, right) versus a high-budget constraint (8000 tokens, left). Each color represents a different category \cite{hou2025thinkprune} of reasoning step. The change in the relative proportions of these steps illustrates how the model strategically allocates its cognitive resources depending on the available budget.}
    \label{fig:behavior_analysis}
\end{figure}

Finally, to answer RQ3, we analyze how BARD adapts its reasoning process. Figure~\ref{fig:behavior_analysis} visualizes the distribution of reasoning steps across different reasoning budgets, contrasting the model's behavior at a low budget (500 tokens) with a high budget (8000 tokens). The results reveal a profound, strategic shift in the model's cognitive approach.

At a high budget, the model adopts an exploratory and verification-driven strategy. It engages in broad, multi-path exploration, akin to a breadth-first search, where it formulates and compares alternative solution paths ("Method 1... Method 2... We find Method 1 is superior..."). More strikingly, it allocates a significant portion of its budget to rigorous self-correction and verification. This manifests as meticulous step-by-step reviews ("Let me re-check this substitution..."), critical examination of its own logical chain for potential flaws ("Is this assumption valid?"), and the ability to correct its trajectory upon discovering an error. The goal shifts from merely finding an answer to constructing a robust, unassailable argument, complete with discussions of boundary conditions and potential pitfalls.

In stark contrast, under a tight budget, the model switches to a core-logic-first, efficiency-driven approach. Its thought process becomes highly focused, prioritizing the most direct path to a solution: problem decomposition, core derivation, and answer generation. It actively "prunes" non-essential cognitive activities. Multi-path exploration is abandoned, and the crucial, yet costly, process of self-correction is almost entirely omitted. The model forgoes intermediate checks ("Was my last step correct?") and minimizes explanations, retaining only the bare-minimum steps required to derive the final answer. This stark difference in behavior confirms that BARD learns not just to shorten its output, but to fundamentally alter its reasoning strategy based on the available computational resources.









\section{Conclusion}
We presented BARD (Budget-Aware Reasoning Distillation), a unified framework that transfers reasoning abilities while enabling fine-grained control over reasoning length. By integrating a user-specified thinking budget into both contrastive Supervised Fine-Tuning and Reinforcement Learning with a multiplicative reward, BARD jointly optimizes reasoning accuracy and budget fidelity.

Experiments on AIME24, AIME25, and GPQA show that BARD achieves superior performance and precise budget control compared to standard distillation and truncation baselines. Moreover, analysis reveals that BARD adapts its reasoning strategy under different budgets—prioritizing concise, goal-oriented reasoning when resources are limited, and engaging in exploration and verification when budgets expand.

In summary, BARD provides an effective path toward resource-efficient reasoning models, offering controllable and adaptive reasoning suitable for real-world deployment.


\bibliographystyle{unsrt}  
\bibliography{references}  

\appendix 

\section{Prompt for Thought Process Compression}
\label{app:prompt} 

This appendix details the prompt used to instruct the LLM to compress its thought process. The placeholders \texttt{\{query\}}, \texttt{\{think\}}, \texttt{\{answer\}}, and \texttt{\{think\_len\}} were dynamically filled during the process.

\begin{verbatim}
## Task Description

You are an expert in compressing "Thought Processes." Please compress the
provided "Thought Process" according to the following requirements:

1. Refer to the input information below (the related Question, Thought
   Process, and Answer). You must analyze the relationship between the
   Question and the Answer, and compress the Thought Process to a
   specified length. It is crucial that you do not alter the original style
   or meaning of the thought process. The compressed thought process must
   serve as a logical bridge between the Question and the Answer, ensuring
   coherence.

2. While compressing the Thought Process to the specified token limit,
   avoid excessive compression. Strive to retain the most critical content
   of the thought process.

3. The first sentence of the original Thought Process must remain
   unchanged.


## Input Information

### Question
{query}

### Thought Process
{think}

### Answer
{answer}


## Output Requirements

1. Compressed Thought Process Length: {think_len} tokens

2. Output format for the compressed thought process:
   Compressed Thought Process:(Ensure no other content follows)
\end{verbatim}
\end{document}